\documentclass{article}

\usepackage[final]{neurips_2020}

\usepackage[utf8]{inputenc} 
\usepackage[T1]{fontenc}    
\usepackage{hyperref}       
\usepackage{url}            
\usepackage{booktabs}       
\usepackage{amsfonts}       
\usepackage{nicefrac}       
\usepackage{microtype}      
\usepackage[ruled,vlined]{algorithm2e}
\usepackage{float}
\usepackage{graphicx}

\title{Limitation Learning: Catching Adverse Dialog with GAIL}
  
\author{%
  Noah Kasmanoff\\
  Center for Data Science\\
  New York University\\
  New York, NY 10011 \\
  \texttt{nsk367@nyu.edu} \\
   \And
   Rahul Zalkikar\\
  Center for Data Science\\
  New York University\\
  New York, NY 10011 \\
  \texttt{rz1567@nyu.edu} \\
}

\begin{document}

\maketitle

\begin{abstract}

Imitation learning is a proven method for creating a policy in the absence of rewards, by leveraging expert demonstrations. In this work, we apply imitation learning to conversation. In doing so, we recover a policy capable of \textit{talking} to a user given a prompt (input state), and a discriminator capable of classifying between expert and synthetic conversation. While our policy is effective, we recover results from our discriminator that indicate the limitations of dialog models. We argue that this technique can be used to identify adverse behavior of arbitrary data models common for dialog oriented tasks.

\end{abstract}

\section{Introduction}

The rising popularity of deep learning owes much credit to the development of algorithms with millions to billions of parameters and as many data-points to train. One sub-field that has especially flourished in this respect is natural language processing (NLP). With far-reaching applications, NLP has seen much development due to specialized applications of large models trained on even larger bodies of text. With fine-tuning capabilities for specific tasks, these models are considered nearly ready for real-world applications such as company chat-bots. Even with proper fine-tuning, there is still a chance that these models produce adverse and harmful responses\footnote{\href{https://artificialintelligence-news.com/2020/10/28/medical-chatbot-openai-gpt3-patient-kill-themselves/}{https://artificialintelligence-news.com/2020/10/28/medical-chatbot-openai-gpt3-patient-kill-themselves/}}. 

In this work, we use generative adversarial imitation learning (GAIL) on a simple dialog model and dataset to tease out these adverse actions. By demonstrating the limits of language models in this setting, we next describe how this method is easily transferable to large language models as a potential method to use for discovering and mitigating harmful actions.

\textbf{Related work}. \cite{li2018dialogue} expanded on the adversarial dialogue generation method introduced by \cite{li2017adversarial} to a new model, DG-AIL, that incorporates an entropy regularization term to the generation objective function. To extend on this work, we employ generative adversarial imitation learning (GAIL) to produce a proxy for the reward function present in conversations from the Cornell Movie Dialog Corpus. We apply imitation learning to craft coherent replies to the input utterance.

The proxy reward function can be used to probe the recovered reward signal. If incoherent or toxic utterances are assigned high reward, we flag these instances as possible adversarial cases for future studies. If done with arbitrary models or data, this technique presents an accelerated way to identify the limitations of a dialog system via imitation.

This work is just the beginning. We emphasize that GAIL is a method of imitation learning, not inverse reinforcement learning. This distinction is important in that we cannot recover the true underlying reward function of the system, but instead a proxy based on the optimal imitation policy. We consider the application of more advanced techniques such as guided cost learning \cite{finn2016guided} a worthwhile next step.

To summarize, we make the following contributions\footnote{Code available at \href{https://github.com/nkasmanoff/limitation-learning}{https://github.com/nkasmanoff/limitation-learning}}:

\begin{itemize}

\item{Train GAIL on the Cornell Movie Dialog Corpus.}
\item{Probe the behavior of the recovered reward signal to discover "adversarial actions" that exist in conversation.}
\item{Present steps to apply this method to arbitrary models and data, and a potential way to analyze black-box models.}
\end{itemize}

\section{Data}

For this work, we use conversations from the Cornell-Movie-Dialog corpus\footnote{\href{https://www.cs.cornell.edu/~cristian/Cornell\_Movie-Dialogs\_Corpus.html}{https://www.cs.cornell.edu/~cristian/Cornell\_Movie-Dialogs\_Corpus.html}}, which contains over 220 thousand conversations and 304k utterances between movie characters in films of various genre. In this dialog environment, we define states and actions as token sequences $<tok_1, tok_2,..., tok_t>$ that form an utterance. We use pre-computed embeddings to project these sequences into an intermediate representation to form padded, vectorized forms of state-action pairs in dialog, while limiting state contexts to a max of 1 or 2 conversation turns. For example, we accept into the data a dialog sequence $s_0, s_1, s_2$, where $s_i = <tok_1, tok_2,..., tok_t>$, and $t$ is the number of padded tokens, to form state/action pairs $[f(s_0),f(s_1)], [f(s_1),f(s_2)], [(f(s_0),f(s_1)),f(s_2)]$ where $f$ is our mapping from a sequence of tokens to a matrix of embedding vectors. 

\textbf{Experimental settings}. The reinforcement learning dialog objective is to maximize an expected reward over a conversation. In our experiments we use a max of 1 conversation turn for state-action pairs. To further simplify things, we impose a max state length of 5 tokens and filter states where every token is not assigned a corresponding word embedding.

\textbf{Word Embeddings}
We pre-process the restricted data with the spaCy\footnote{\href{https://spacy.io/api}{https://spacy.io/api}} Python module for natural language processing and its pre-trained model, en\_core\_web\_lg\footnote{\href{https://github.com/explosion/spacy-models/releases//tag/en_core_web_lg-2.3.1}{https://github.com/explosion/spacy-models/releases//tag/en\_core\_web\_lg-2.3.1}}. We leverage this large vector model specifically for its pre-trained named entity recognizer and novel tokenization algorithm. en\_core\_web\_lg was trained on OntoNotes and Glove Common crawl datasets. Expanding on the pre-processing method from \cite{serban2016processing}, all persons, cardinal numbers, dates, and other entities (as defined by en\_core\_web\_lg) are replaced with special tokens <person>, <cardinal>, <date>, etc. respectively. We add tokenization exceptions to en\_core\_web\_lg to preserve these tokens.

To form our word embeddings we initialize a gensim\footnote{\href{https://pypi.org/project/gensim/}{https://pypi.org/project/gensim/}} Word2Vec model with our corpus vocabulary. We load the multi-purpose Google News Vectors\footnote{\href{https://code.google.com/archive/p/word2vec/}{https://code.google.com/archive/p/word2vec/}} which contain embeddings for 3 million words and phrases. We prune our vector space by loading only the embeddings for which there are already words in our vocabulary. We modify the default gensim training behavior by supplying 1.0 for the $lockf$ parameter so that all tokens in our vocabulary, either initialized by a Google News embedding or unique to our vocabulary (ex. <person>), are adjusted during training. We note that this was an experimental process and that we manually examined the tokens in our final model along with the most similar tokens to our special entities based on their vector's $l_2$-normed cosine similarities. 

The final model contained $l_2$-normalized, 300 dimensional embeddings for 1929 tokens which appeared a minimum of 5 times in the corpus. Tokens included punctuation and lowercase alphanumeric and special characters. Top tokens by count include: "." (41324), "?" (27176), "you" (9447), "<person>" (9056), "what" (8112), "i" (7164), "!" (7094), "..." (6689), "," (6063), "'s" (4781), "it" (4287) and "no" (3334).

\section{Methods}

\textbf{Preliminaries}. We build our environment as a Markov Decision Process (MDP), defined by a tuple $(S,A,\tau, r, \gamma)$ where $S$ and $A$ are the state and action space respectively, $\tau$ is the transition probability, such that $\tau(s,a,s')$ is the probability of transitioning from state $s$ under action $a$. $r(s,a)$ is the immediate reward after taking action $a$ while in state $s$. Although we define the discount factor $\gamma$ for completeness, our environment does not require it. Trajectories are simply pairs of one state and one action. The optimal policy we seek is a mapping $\pi$ which accepts states $s$ to actions $a$ with probability $\pi(a | s)$. 

\textbf{Model}. For the policy, we employ a sequence to sequence (Seq2Seq) model \cite{luong2015attention} built from a 2-layer, bi-directional GRU encoder with 128 hidden units and a subsequent linear layer, and a 2-layer GRU decoder with a subsequent linear layer and an output size equal to that of our Word2Vec (w2v) vocabulary. Our embedding layer is initialized to our pre-trained w2v model and frozen to updates.

$<sos>$ and $<eos>$ are special tokens initialized to random $l_2$-normed vectors in our w2v vocabulary. They represent the start and end of a sequence of tokens. An attention mechanism is incorporated into the decoder and we use packed padded sequences and masking. Each utterance is padded to a max length of 5 tokens. 

\textbf{Generative Adversarial Imitation Learning (GAIL)}. GAIL casts the objective of imitation learning into a min max optimization problem analogous to generative adversarial networks (GANs) \cite{ho2016generative}, \cite{goodfellow2014generative}. Similar to GANs, a discriminator $D$ is used to distinguish state-action pairs coming from the expert demonstrations $\pi_E$, and that of the learned policy $\pi$. By optimizing $D$, we obtain a non-linear approximator that surpasses behavioral cloning, which is limited by its inherent i.i.d assumptions. 

\begin{figure}[H]
  \includegraphics[width=8cm]{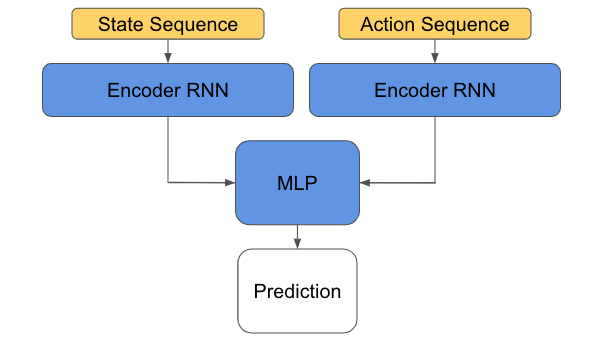}
  \centering
  \caption{Flow diagram of the discriminator architecture. Two sequences are accepted by the two encoder modules, and an output prediction assigns a probability between $0$ (expert action) and $1$ (learner action) to characterize trajectories.}
     \label{figure:discrim}

\end{figure}

The discriminator (Fig. \ref{figure:discrim}) attempts to minimize the loss function $E_{\pi}[log(D(s,a))] + E_{\pi_E}[log(1-D(s,a))]$ for separating policy and expert actions for a given a state input. Concurrently, the learned policy $\pi$ attempts to maximize the proxy for the reward function, $ r(s,a) = -log(D(s,a))$. As this reward increases, the discriminator has a harder time classifying between the two input classes. This policy objective is formulated as $E_{\tau}[\nabla_{\theta} log(\pi_{\theta}(a | s) r(s,a)]  $.

The policy is trained through standard model-free reinforcement learning techniques. When successful, the policy is able to "trick" the discriminator into classifying its trajectory as that of the expert. As the policy improves, the discriminator gets better at distinguishing trajectories, and the game continues. 

\textbf{Training}.  We primed our networks by first pre-training the policy network with behavioral cloning. Doing so provided us with an initialized encoder and sequence to sequence architectures, all of which are  inserted into the networks used for GAIL. The encoders are used for the discriminator, and the sequence to sequence architecture for the policy. 

We perform GAIL with minor changes from its original implementation. We present our adapted algorithm below.  

\begin{algorithm}[H]
\SetAlgoLined
\textbf{Input} Initial policy and discriminator parameters $\theta_0, \omega_0$ , pretrained via behavioral cloning  \;
 \While{\textit{iter} < max\_iter}{
  Sample trajectories $\tau_i$ \texttildelow $\pi_i$\;
  Update discriminator parameters from $\omega_i$ to $\omega_{i+1}$ with the gradient\;
  $$E_{\tau_i}[\nabla_{\omega}log(D_{\omega}(s,a))] + E_{\tau_{E_i}}[log(1-D_{\omega}(s,a))]$$
  
  Update actor parameters from $\theta_i$ to $\theta_{i+1}$ with the gradient\;
  Specifically, take a policy gradient step with \;
  $$E_{\tau_i}[\nabla_{\theta} log(\pi_{\theta}(a | s)) r_{GAIL}(s,a)] $$
  }
\caption{Modified GAIL}
\end{algorithm}

In our case, we use 5 discriminator update steps for every 1 policy update step. Both the agent and discriminator are trained via the Adam optimizer \cite{kingma2017adam} with Pytorch \cite{pytorch} on a NVIDIA V100 GPU on the New York University High Performance Compute Cluster \textit{Prince}.

After training GAIL, we recover a policy capable of response, and discriminator for quantifying responses. 
Our final model was pre-trained for 30 epochs on over 8000 dialogue state pairs and over 16000 sequences, and we observed its convergence to an optimal averaged cross entropy loss and cosine similarity. Next, it was trained using GAIL for 12000 randomly selected (without replacement) states.

\section{Results}

\begin{figure}[H]
\centering
  \includegraphics[width=4cm]{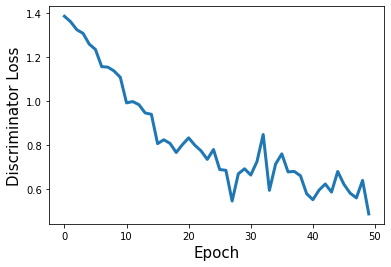}
    \includegraphics[width=4cm]{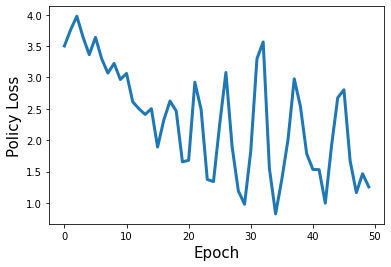}
      \includegraphics[width=4cm]{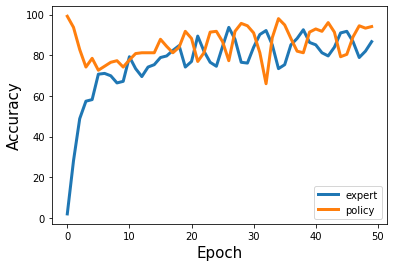}

  \caption{How we can evaluate the training of our adversarial imitation learning. From left to right, we examine the gradual convergence and variation in the magnitude of our discriminator loss, policy objective function, and overall accuracy as a function of state examined. Note that the policy loss is inherently unstable. With the discriminator network improving, the policy loss will "jump" first before declining. }
     \label{figure:GAILStuff}

\end{figure}

We first investigate the stability of GAIL training. To do so we display the discriminator loss, policy objective loss, and discriminator accuracy in Fig. \ref{figure:GAILStuff}. As seen in these diagrams, we observe that our networks steadily improve, and the discriminator effectively distinguishes between expert and policy sentences. While the results were overwhelmingly positive, what we are interested in discovering are cases where the discriminator was fooled. In table \ref{table:goodpolicy}, we look at instances where our agent received the highest reward. Similarly, in table \ref{table:badpolicy}, we explore instances of actions that were deemed worst by the discriminator, or had the lowest rewards. In table \ref{table:policy} we examine how a stochastic agent produces varied responses to the same prompt, and compare that to the actions obtained via behavioral cloning.

From these, we discover that the most frequent cause of either success or failure was the same; when the expert demonstration was a follow-up question to the input state. In the highest reward cases it assigned a word + question mark and in the lowest reward cases it assigned only question marks. This indicates that the discriminator created a reward function which assigned some value to the presence of words and punctuation as important parts of the reply. Even though just question marks paint the same semantic meaning, the discriminator penalized the absence of words. Furthermore, given that the question mark is one of the most common tokens in the corpus, it is no surprise to see it so frequently used by the policy.

\begin{figure}[H]
\centering
  \includegraphics[width=6.5cm]{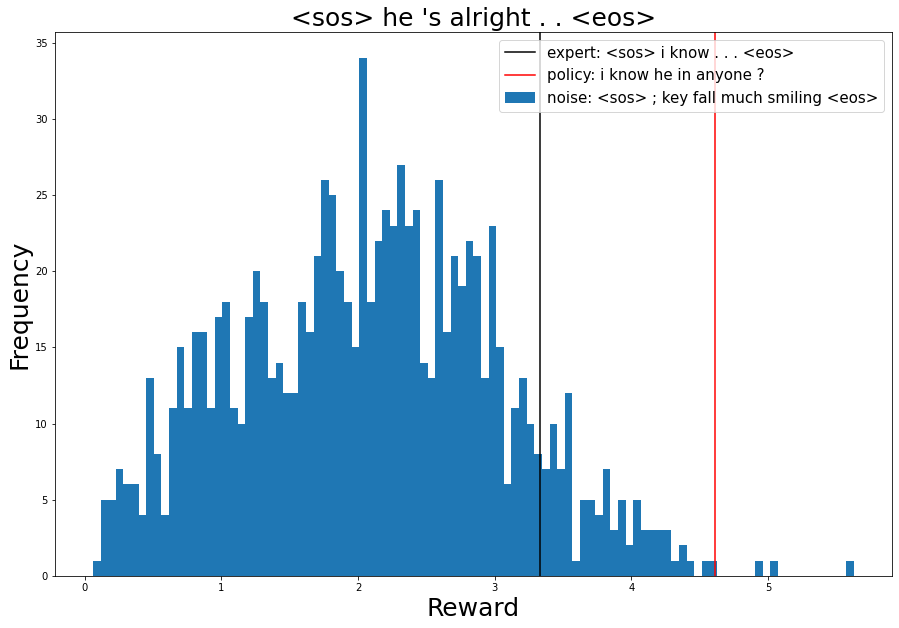}
    \includegraphics[width=6.5cm]{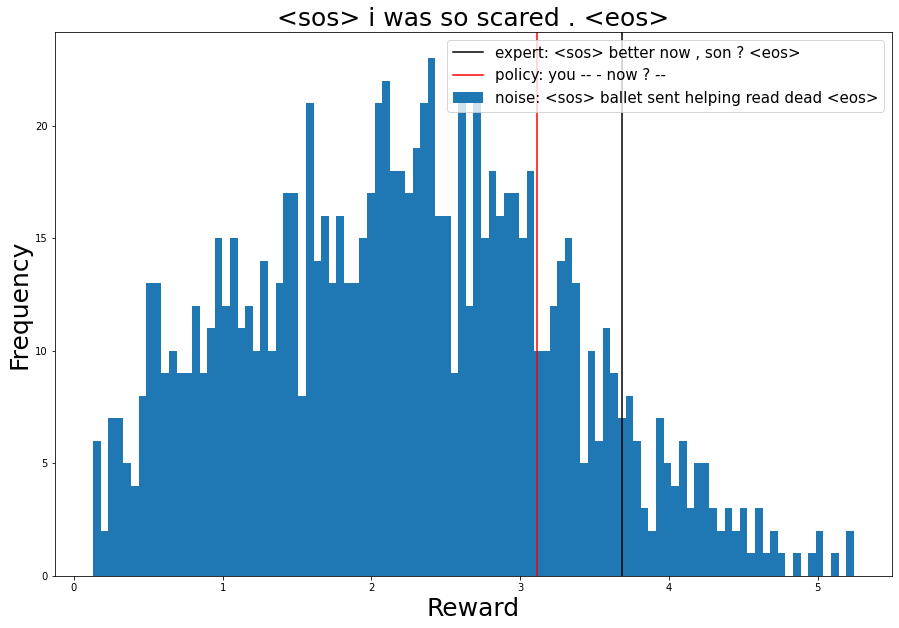}

  \centering
  \caption{Histogram of reward signal of randomly sampled inputs. We find that while most noise does not fool the discriminator, the random actions with high reward can be considered adversarial. The legend provides the highest reward action from the expert, policy, and noise.}
     \label{figure:rewarddist}

\end{figure}

We next probe the reward function with random actions. In Fig. \ref{figure:rewarddist}, after inputting 5 random sequence tokens, we calculate the reward of that action and compare over many such instances the distribution of this reward signal compared to the expert and policy rewards. It is encouraging to see the policy and expert action reward signals to be similar to one another when also examining the raw tokens of each. However, what we discover, and consider crucial to understanding conversational models, are the few sequences which, despite having nonsensical inputs, produce high reward. We observe the potential issues that could arise in dialog models if these were to be produced in an applied setting.

\section{Conclusion}

We have shown, for the \textit{first} time, how to use generative adversarial imitation learning to characterize the sensitivity of dialog generation language models for producing adverse and harmful replies.

Imitation and inverse reinforcement learning are one way to catch before deployment, harmful utterances. The policy is trained by fooling a discriminator network using model free reinforcement learning. With sufficient training, the reward function provides insight into what actions might have been accidentally produced if this network was trained differently. This work will benefit the deployment of conversational AI by introducing a novel way of poking holes in the model's acceptable action space.
 
This work has made use of a specific kind of conversation, those found in film, but could be applied to any conversational data. As a future direction, we are especially interested in seeing this technique used on synthetic dialog produced by large language models such as GPT3 \cite{brown2020language} for conversations with humans. In this scenario, we could discover potential cases of adverse actions by these large models using the methods described in this work. To do this, one simply need replace the model used for the policy and its corresponding encoder for the reward generating discriminator. With minimal knowledge and insight into the underlying models we seek to probe, we can recover instances of sequences which may be erroneously produced if left unchecked.

\section*{Figures and Tables}

\begin{table}[H]
  \caption{Top 10 most similar tokens to "<person>"}
  \centering
  \begin{tabular}{lll}
    \toprule
    \cmidrule(r){1-2}
    Token & $l_2$-normalized cosine similarity \\
    \midrule
    morning & 0.5010  \\
    hello & 0.4668 \\
    monsieur & 0.4654 \\
    madame & 0.4624 \\
    goodbye & 0.4482 \\
    goodnight & 0.4467 \\
    babe & 0.4452 \\
    mrs. & 0.4416 \\
    <org> & 0.4399 \\
    daddy & 0.4286 \\
    \bottomrule
  \end{tabular}
    \label{table:tokensims}
    
\end{table}

\begin{table}[H]
  \caption{Behavioral Cloning Pre-Training (PT) Policy vs. GAIL Policy (\textbf{**} had higher reward) }
  \centering
  \begin{tabular}{llll}
    \toprule
    \cmidrule(r){1-2}
    State & Expert & PT Policy & GAIL Policy\\
    \midrule
    how 's everything ? . & everything is everything . & it 's everything treat them & 
    \begin{tabular}[t]{@{}l@{}} it 's everything family . \\ it 's everything treat . \ \ \textbf{**} \end{tabular} \\ \cline{1-4}
    good <time> , doctor . & good <time> ! & good <time> ! ! ! & 
    \begin{tabular}[t]{@{}l@{}} you captain . \ \ \textbf{**} \\ day ! \end{tabular} \\ \cline{1-4}
    <time> <time> to go ! & yes . & yes -- yes ? ? & 
    \begin{tabular}[t]{@{}l@{}}you after ? \\ you yes ? \ \ \textbf{**} \end{tabular} \\ 
    \bottomrule
  \end{tabular}
\label{table:policy}
\end{table}

\begin{table}[H]
  \caption{GAIL Optimal Policy vs. Expert Responses: Highest Reward Actions}
  
  \centering
  \begin{tabular}{lll}
    \toprule
    \cmidrule(r){1-2}
    State     & Policy     & Expert \\
    \midrule
    i was just wondering & ok what ? she good ?   & wondering what ?     \\
    alone at last     & what for <gpe>  & now where were we ?     \\
     the other car !     &  <person>    & dr. <person> ! \\
     
       speak <language> !     & no said speak <norp> ? !     & you said speak <norp> ! \\
    \bottomrule
  \end{tabular}
      \label{table:goodpolicy}

\end{table}

\begin{table}[H]
  \caption{GAIL Optimal Policy vs. Expert Responses: Lowest Reward Actions}
  
  \centering
  \begin{tabular}{lll}
    \toprule
    \cmidrule(r){1-2}
    State     & Policy     & Expert \\
    \midrule
    turn around ! & ? ? ?  & what ?     \\
   can you see anything ?     &  wo , - & what is it ?    \\
    she bought it     & ? ? ?       & why ? \\
    \bottomrule
  \end{tabular}
      \label{table:badpolicy}

\end{table}


\bibliographystyle{unsrt}
\bibliography{references}

\begin{thebibliography}{10}

\bibitem{li2018dialogue}
Ziming Li, Julia Kiseleva, and Maarten de~Rijke.
\newblock Dialogue generation: From imitation learning to inverse reinforcement learning, 2018.

\bibitem{li2017adversarial}
Jiwei Li, Will Monroe, Tianlin Shi, Sébastien Jean, Alan Ritter, and Dan Jurafsky.
\newblock Adversarial learning for neural dialogue generation, 2017.

\bibitem{finn2016guided}
Chelsea Finn, Sergey Levine, and Pieter Abbeel.
\newblock Guided cost learning: Deep inverse optimal control via policy optimization, 2016.

\bibitem{serban2016processing}
Iulian~V. Serban, Alessandro Sordoni, Yoshua Bengio, Aaron Courville, and Joelle Pineau.
\newblock Building end-to-end dialogue systems using generative hierarchical neural network models, 2016.

\bibitem{luong2015attention}
Minh-Thang Luong, Hieu Pham, and Christopher~D. Manning.
\newblock Effective approaches to attention-based neural machine translation, 2015.

\bibitem{ho2016generative}
Jonathan Ho and Stefano Ermon.
\newblock Generative adversarial imitation learning, 2016.

\bibitem{goodfellow2014generative}
Ian~J. Goodfellow, Jean Pouget-Abadie, Mehdi Mirza, Bing Xu, David Warde-Farley, Sherjil Ozair, Aaron Courville, and Yoshua Bengio.
\newblock Generative adversarial networks, 2014.

\bibitem{kingma2017adam}
Diederik~P. Kingma and Jimmy Ba.
\newblock Adam: A method for stochastic optimization, 2017.

\bibitem{pytorch}
Adam Paszke, Sam Gross, Francisco Massa, Adam Lerer, James Bradbury, Gregory Chanan, Trevor Killeen, Zeming Lin, Natalia Gimelshein, Luca Antiga, Alban Desmaison, Andreas Kopf, Edward Yang, Zachary DeVito, Martin Raison, Alykhan Tejani, Sasank Chilamkurthy, Benoit Steiner, Lu~Fang, Junjie Bai, and Soumith Chintala.
\newblock Pytorch: An imperative style, high-performance deep learning library.
\newblock In H.~Wallach, H.~Larochelle, A.~Beygelzimer, F.~d\textquotesingle Alch\'{e}-Buc, E.~Fox, and R.~Garnett, editors, {\em Advances in Neural Information Processing Systems 32}, pages 8024--8035. Curran Associates, Inc., 2019.

\bibitem{brown2020language}
Tom~B. Brown, Benjamin Mann, Nick Ryder, Melanie Subbiah, Jared Kaplan, Prafulla Dhariwal, Arvind Neelakantan, Pranav Shyam, Girish Sastry, Amanda Askell, Sandhini Agarwal, Ariel Herbert-Voss, Gretchen Krueger, Tom Henighan, Rewon Child, Aditya Ramesh, Daniel~M. Ziegler, Jeffrey Wu, Clemens Winter, Christopher Hesse, Mark Chen, Eric Sigler, Mateusz Litwin, Scott Gray, Benjamin Chess, Jack Clark, Christopher Berner, Sam McCandlish, Alec Radford, Ilya Sutskever, and Dario Amodei.
\newblock Language models are few-shot learners, 2020.

\end{thebibliography}

\end{document}